# Binary and nonbinary description of hypointensity in human brain MR images

Xiaojing Chen

A thesis submitted to the Department of Mathematics and Natural Sciences in partial fulfilment of the requirements for the degree of

Master of Science in Computer Science

Supervisor: Dr. Michael S. Lew

**Leiden University**

Authors' address    X. Chen                              tracy_chen69@hotmail.com

| Title: | Binary and nonbinary description of hypointensity in human brain MR images |
|---|---|
| Author: | X. Chen |

**Abstract:** Accumulating evidence has shown that iron is involved in the mechanism underlying many neurodegenerative diseases, such as Alzheimer's disease, Parkinson's disease and Huntington's disease. Abnormal (higher) iron accumulation has been detected in the brains of most neurodegenerative patients, especially in the basal ganglia region. Presence of iron leads to changes in MR signal in both magnitude and phase. Accordingly, tissues with high iron concentration appear hypo-intense (darker than usual) in MR contrasts. In this report, we proposed an improved binary hypointensity description and a novel nonbinary hypointensity description based on principle components analysis. Moreover, Kendall's rank correlation coefficient was used to compare the complementary and redundant information provided by the two methods in order to better understand the individual descriptions of iron accumulation in the brain.

**Conclusions:** We presented novel methods to describe hypointensity of human brain in MR images. Besides that, several hypointensity features were defined in both binary and nonbinary descriptions. Analysis of relations of these features revealed high within-feature correlations, while the between-feature correlations were at medium strength. After evaluation of the performances of the proposed methods with the ground truth of SWI dataset, we concluded that, overall, nonbinary hypointensity description is better than binary hypointensity description and the highest accuracy was achieved when using nonbinary method with spatially balanced sampling. Furthermore, comparison results from Golden dataset confirmed the reliability and robustness of both methods. The possible future work may further involve the investigation of the optimum ROI localization and evaluation of improved descriptions based on a more objective ground truth, as well as extending the current description methods onto 3D.

# Acknowledgement

I would like to express my gratitude to the Leiden Institute of Advanced Computer Science at Leiden University and Video Processing and Analysis of Philips Research for giving me the permission to complete this thesis to fulfill the master program.

Furthermore, I'm deeply grateful to my supervisor at Philips, Dr. Devrim Unay, whose stimulating suggestions and generous encouragement helped me in all phases of research and development and in writing this thesis. And I wish to extend my thanks to my advisor at Leiden University, Dr. Michael S. Lew, for his valuable advices and instructions on my master graduation project.

Thanks are also due to Ahmet Ekin and Radu Jasinschi, for their kind helps and suggestions on my experiments. And I give my special thanks to my dear colleagues who provided me with such a delightful working environment and never failed to encourage and assist me.

At last but not least, I owe many loving thanks to my family for their never-ending support and infinite belief in me all the way from the very beginning of my postgraduate study.

# Contents

# List of figures

# List of tables

# 1.    Introduction

## 1.1.    Age-related  neurodegenerative diseases and iron accumulation

As Population aging – explained as a shift in the distribution of a country's population towards greater ages – has become a more and more pervasive and severe issue all over the world, age-related medical and healthcare problems are increasing correspondingly. Age is considered as one of the primary risk factors for certain neurological diseases, such as Alzheimer's disease, Parkinson's disease and Huntington's disease. For example, in the U. S. alone, 2.5 million people are suffering from Alzheimer's diseases and more than 300,000 new cases are diagnosed per year. More importantly, the Alzheimer's Association expects those numbers to triple in the next 20 years[1]. However, currently neither detection nor diagnosis of most of the neurodegenerative diseases is a fully solved problem due to either lack of sufficient understanding of some diseases or the inter/intra-observer variability of the diagnoses[2].

At the molecular level, all neurodegenerative diseases are associated with iron accumulation in deep gray matter structures of human brain. It is well known that iron is required to maintain the brain's optimal functioning, and therefore is essential for life. In 1958, Hallgren and Sourander[3] performed some of the earliest work on characterizing brain iron. To date, post-mortem and *in vivo* studies have demonstrated that in normal individuals, iron levels increase with age in subcortical and some cortical gray matter regions[4][5][6][7]. In the globus pallidus, red nucleus, substantia nigra, and dentate nucleus, iron increases rapidly from birth until the end of the second decade, plateaus for several years, and then shows another milder increase after age 60. Iron increases more slowly in the putamen and caudate, levelling off in the fifth or sixth decade[8]. However, there is accumulating evidence that iron is involved in the mechanisms underlying many neurodegenerative diseases[9][10]. Abnormal iron deposition in the basal ganglia has been detected in most age-related neurodegenerative diseases.

## 1.2.    MRI background

Magnetic resonance imaging (MRI), is primarily a medical imaging technique most commonly used in radiology to visualize the structure and function of the body. It provides detailed images of the body in any plane. MRI provides much better contrast between the different soft tissues of the body than computed tomography (CT) does, making it especially useful in neurological (brain), musculoskeletal, cardiovascular, and oncological (cancer) imaging. Unlike CT, it uses no ionizing radiation, but uses a powerful magnetic field to align the nuclear magnetization of (usually) hydrogen atoms in water in the body. Radio frequency fields are used to systematically alter the alignment of this magnetization, causing the hydrogen nuclei to produce a rotating magnetic field detectable by the scanner. This signal can be manipulated by additional

                                                                                    

magnetic fields to build up enough information to construct an image of the body. While CT provides good spatial resolution (the ability to distinguish two structures an arbitrarily small distance from each other as separate), MRI provides comparable resolution with far better contrast resolution (the ability to distinguish the differences between two arbitrarily similar but not identical tissues).

Image contrasts (T1, T2 and T2 * ) are created by differences in the strength of the nuclear magnetic resonance signal recovered from different locations within the sample[11]. Contrast in most MR images is actually a mixture of all these effects, but careful design of the imaging pulse sequence allows one contrast mechanism to be emphasized while the others are minimized. The ability to choose different contrast mechanisms gives MRI tremendous flexibility. In the brain, T1-weighting causes the nerve connections of white matter to appear white, and the congregations of neurons of gray matter to appear gray, while cerebrospinal fluid (CSF) appears dark, as shown in Figure 1(a). The contrast of white matter, gray matter and cerebrospinal fluid is reversed using T2 or T2 * imaging, whereas proton-density-weighted imaging provides little contrast in healthy subjects, as shown in Figure 1(b). Additionally, functional parameters such as cerebral blood flow (CBF), cerebral blood volume (CBV) or blood oxygenation can affect T1, T2 and T2 *.

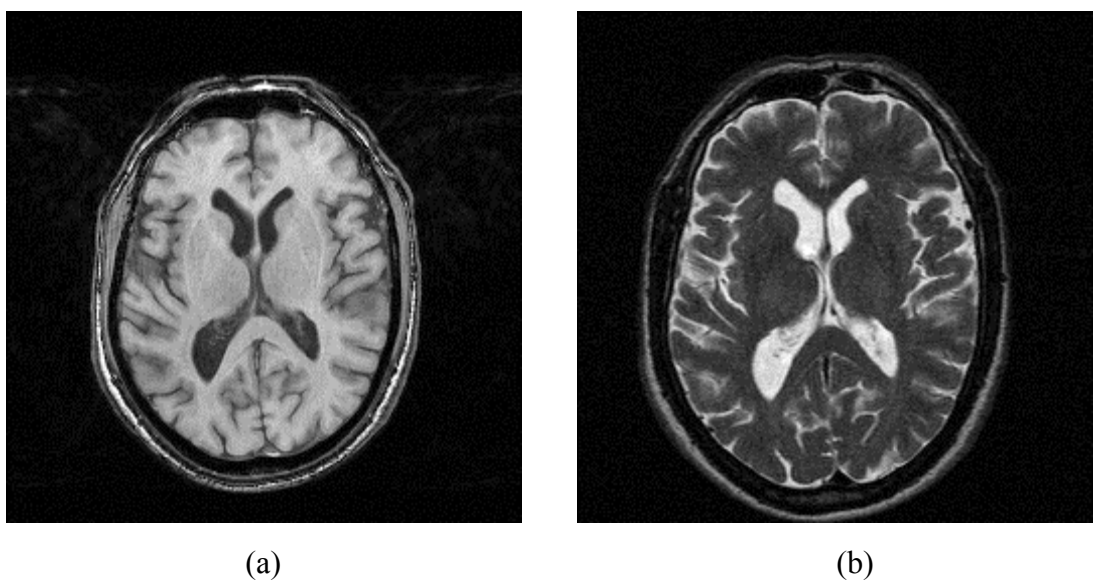

(a)                                    (b)

Figure 1: T1 and T2 brain MRI scans. (a) is a Noncontrast axial T1-weighted MR image and (b) is its corresponding T2-weighted MR image.

Susceptibility-weighted imaging (SWI) is a novel magnetic resonance (MR) technique that exploits the magnetic susceptibility differences of various tissues, such as iron, blood and calcification. SWI consists of using both magnitude and phase images from a high-resolution, three-dimensional fully velocity-compensated gradient echo sequence[12]. Phase mask is created from the MR phase images, and multiplying these with magnitude

 

images increase the conspicuity of the smaller veins and other sources of susceptibility effects[13].

It is well known that presence of iron leads to changes in MR signal in both magnitude and phase. Accordingly, tissues with high iron concentration appear hypo-intense (darker than usual) in MR contrasts, such as T2 and T2*, which focus mainly on the signal magnitude. Due to presence of a phase difference between tissues with iron and those without, SWI exhibits enhanced signal difference, and therefore is promising in non-invasive quantification of brain iron by computer-aided tools.

However, there are also some inevitable problems with MR imaging. In order to compare certain features in several MR images, registration is needed to align all images so that common features can overlap making differences emphasized[14]. Nevertheless, currently no perfect registration method exists that can assure no error occurs during the geometric transformation procedure. Also spatially misalignment may appear locally even when global registration is applied. Another problem is intensity variation in MR images. On one hand, the variation range of pixel values in one image may differ from another, for example, typically the range of pixel values in a grayscale image is from 0 to 255, while in a MR image this range may change to 0 to 1200. On the other hand, brightness variations of the same tissue may appear across the image. Such intensity inhomogeneities often arise from what is usually called the bias field, and are due to inhomogeneities in the static magnetic[15]. In severe cases this can result in ineffective segmentation and registration. Furthermore, susceptibility-weighted imaging suffers more from these issues.

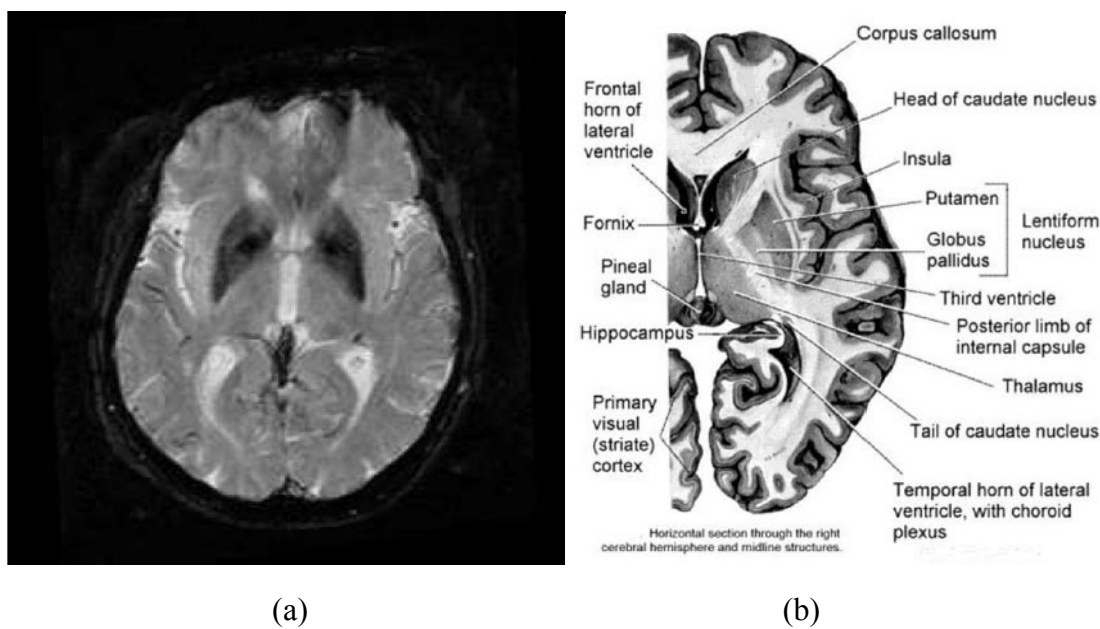

(a)                              (b)

Figure 2: SWI image of brain and a corresponding atlas.

Figure 2(a) demonstrates the iron accumulation effect on a susceptibility-weighted image (SWI) with significantly darker regions in the globus pallidus and the putamen. A corresponding atlas for approximately the same slice is also shown on the right. The structures relevant for iron accumulation include putamen, globus pallidus, and caudate nucleus, which form the basal ganglia responsible for motor control, cognition, emotions and learning. Any abnormality in this region may lead to devastating disorders, including Alzheimer's disease, Parkinon's disease and Huntington's disease.

## 1.3.　Motivation

Currently, there's a growing research interest in the (early) detection, understanding, and diagnosis of neurodegenerative diseases from various disciplines [22][23][24]. The known accumulation of iron in the basal ganglia and the manifested intensity reduction from this area with conventional T2 weighted imaging provide possibility of using MR images to assess the severity of neurodegenerative diseases. Various ways of noninvasive iron quantification methods have been explored during these two decades. Previous methods were mostly based on measuring relaxation time of T2 images; currently the application of high field strength and MFC imaging has gained more convincing results [25][26]. Besides the relaxation time, T2 hypointensity has also been considered as a marker of iron deposition [23]. However, the spatial information contained in hypointensity has not been employed yet.

Our general aim is to describe brain MR images based on similarity of iron deposition in the basal ganglia structure. Actually, basal ganglia structure is composed of several parts, but we are mainly focused on the putamen and globus pallidus, referred as the region of interest (ROI) in our application, as shown in Figure 2(b). In a previous study, a search and retrieval system for brain MR databases was introduced, where iron accumulation was quantified using a binary hypointensity description approach and it was shown that brain iron accumulation shape provides additional information compared to the shape-intensity features[2]. In this paper, besides a more robust binary description, we also proposed a novel nonbinary description of hypointensity based on principle component analysis and studied the complementary and similar information provided by the two descriptions. Furthermore, performances of both binary and nonbinary hypointensity descriptions were estimated in two dataset by comparing with the ground truth.

In the following section, we first introduce our dataset and ground truth as well as the binary and nonbinary hypointensity descriptions we proposed. Section 3 presents how the hypointensity description algorithms are developed by using more representative input data and optimizing the algorithms themselves. In Section 4, we define the features of both binary and nonbinary hypointensity description and analyze these features with Kendall's rank correlation coefficient. Section 5 provides estimation of the binary and

 

nonbinary hypointensity description methods. Finally, in Section 6 we conclude this report and give our recommendations.

# 2.    Hypointensity description

In this section, we first give a brief introduction on the dataset and the ground truth we employed, then explain the general idea of the hypointensity description methods we proposed. More detailed steps for implementation are included in the next section.

## 2.1.    Dataset and ground truth

The dataset includes 40 subjects with atherosclerotic risk factors. MRI was performed on a Philips Intera 1.5T whole body scanner at Leiden University Medical Center. Resulting volumetric data was composed of 22 slices with $256 \times 256$ pixels and $0.85 \times 0.85 \times 6$ $mm^3$ resolution for SWI. From the same location we selected one susceptibility MR image for one patient's MR scan, which depicted relatively explicit outline of the basal ganglia, and totally 40 susceptibility MR images composed our dataset.

Due to the problems that are involved in MR imaging, pre-processing of all selected images was required. Firstly uniform variation range of pixel values ([0, 255]) was applied on all images through normalization by using:

$$I_{norm} = \frac{I - U_{Min}}{Scale} + I_{Min} \qquad\qquad (1)$$

where $I$ is the matrix of all pixel values in one image, $U_{Min}$ is the minimal value in the uniform variation range, $Scale$ is the scaling between current range and the uniform one, and $I_{Min}$ is the minimal value in current variation range. Then we eliminated the images that had particularly severe bias field problem, for example, abnormal bright/dark areas appeared in the basal ganglia region. Actually, there are several intensity correction techniques especially designed for MR images. However, most of them are aim at traditional contrasts, such as T1 and T2. As our aim is not to propose or assess intensity correction algorithms for SWI, we removed those strongly defected images that would affect the application of our methods badly. After the first selection the size of our dataset was reduced to 37. Finally, registration was carried out on all images to overlap the basal ganglia region.

A ground truth is needed in order to evaluate the performance of hypointensity description that we proposed. However, absolutely objective ground truth of brain iron load is too difficult to obtain, since it needs manual annotation of iron at pixel-level on the images. Thus, in order to make estimation of the performance of methods feasible, we just clustered our dataset instead of giving every subject a definite iron load value. Three experts were asked to categorize the database into three clusters: dark, mid and light, according to the darkness of basal ganglia region. Their decisions are made in blindfolded manner. The ground truth is generated based on the majority decision. Since

                                             

there were three experts, at least two of them would give the same label to one subject. Finally, there are 11 subjects in dark cluster and 13 subjects in both mid and light clusters. Figure 3 shows some exemplary susceptibility images for each cluster. Undoubtly, such ground truth still involves subjectiveness because of differences in the experience levels and also the possibility of mental and physical tiredness of the experts.

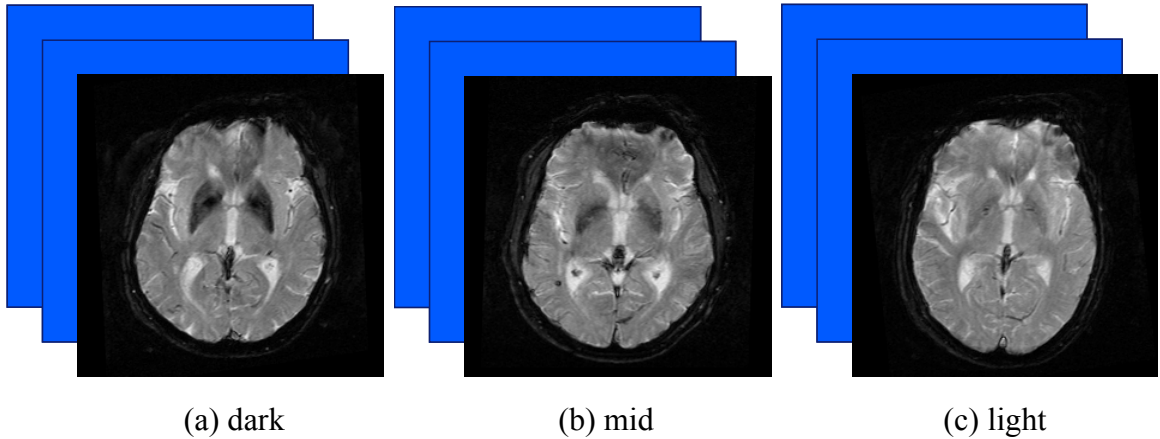

      (a) dark              (b) mid             (c) light

Figure 3: Exemplary susceptibility images for dark, mid and light clusters respetively.

## 2.2. Binary hypointensity description

In susceptibility weighted images (SWI) tissues with higher iron concentration appear hypointense (darker than usual), therefore we proposed an intuitive hypointensity description method in which the percentage of hypointense pixels in an ROI is used to quantify the darkness of basal ganglia region, as explained below.

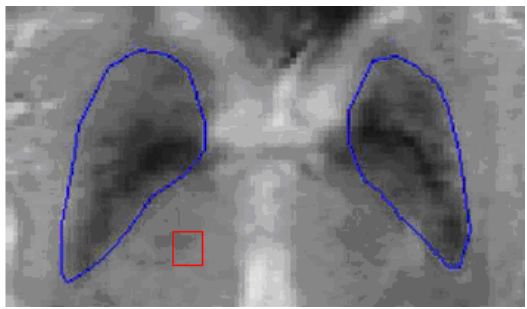

Figure 4: ROI and reference region

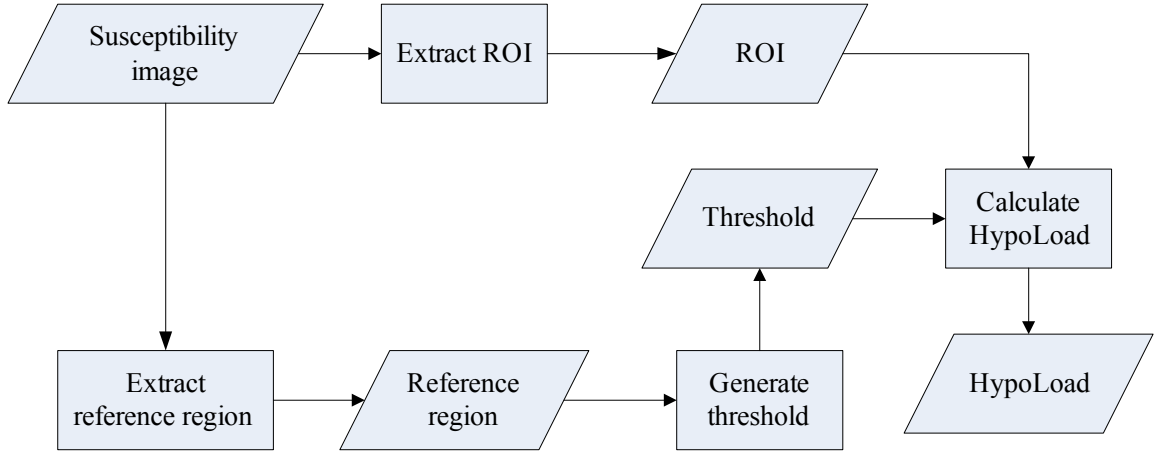

Figure 5: The application process of binary hypointensity description. The rectangles indicate the processing steps and the parallelograms stand for the input or output data, the following figures are of the same structure.

First, pixel values in the ROI of each susceptibility images are extracted. After that, we select a reference region (marked by red rectangle in Figure 4) near the ROI and generate a threshold according to the pixel values in the reference region using Eqn. 2, where $I_{ref}$ is the matrix that contains all pixel values in the reference region.

$$Threshold = mean(I_{ref}) - std(I_{ref}) \qquad (2)$$

Finally, we compare all pixel values in the ROI with the threshold, and the pixels which have values smaller than the threshold will be treated as hypointense. And hypointensity for one subject is defined as the hypointense pixel ratio in the ROI, which we referred to as HypoLoad (Eqn. 3) in our application.

$$HypoLoad = \frac{\sum_{i=1}^{N} S_i}{N} \qquad (3)$$

Where $N$ is the number of pixels in the ROI, and $S_i$ is subject to Eqn. 4, in which $p_i$ is pixel value of pixel $i$ and $T$ is the threshold.

$$Si = \begin{cases} 1, & p_i < T \\ 0, & otherwise \end{cases} \qquad (4)$$

It is easy to notice that this way of defining threshold has one drawback. That is, the selection of reference region is quite subjective with no quantified criteria, which leads to uncertainty of the location and the size of optimal reference region. And this problem will directly affect the robustness of the binary hypointensity description method. Therefore, in the following section we introduced better thresholding approach to solve this problem.

                    

## 2.3. Nonbinary hypointensity description

### 2.3.1. Principle components analysis

Principle component analysis (PCA), also known as Karhunen-Loève transform, is a way of identifying patterns in data and expressing data in such a way as to highlight their similarities and differences[16]. One of the main advantages of PCA is that once these patterns have been found, they can be used to compress the data via vector space transformation. Unlike the binary hypointensity description approach explained above which only gives iron load information, in this nonbinary method based on PCA we are expecting to obtain additional information in the spatial distribution of hypointensity (within the ROI) while eliminating noise in the original data through dimensionality reduction. To give an insight into the nonbinary hypointensity description we proposed based on PCA, we first introduce the PCA algorithm.

Given data X consisting of N images, the data normalization is performed by subtracting the mean vector $m$ from the data. Then the covariance matrix $\Sigma$ of the normalized data ($X - m$) is computed.

$$m = \frac{1}{N} \sum_{i=1}^{N} X_i \qquad (5)$$

$$\Sigma = (X - m)(X - m)^T \qquad (6)$$

Afterwards, the basis functions are obtained by solving the algebraic eigenvalue problem

$$\Lambda = \phi^T \Sigma \phi \qquad (7)$$

where $\phi$ is the eigenvector matrix of $\Sigma$, and $\Lambda$ is the corresponding diagonal matrix of eigenvalues. Consequently, projection of a new data $Y$ to the eigenspace is achieved by

$$G = \Sigma^T (Y - m) \qquad (8)$$

where $G = (g_1, g_2, ..., g_N)$ and $\Sigma = (e_1, e_2, ..., e_N)$ with $g_i$ being a scalar value representing the degree-of-match between the image and the eigenvector $e_i$. Finally, a data (image) $X_j$ can be reconstructed exactly by weighted sum of the eigenvectors.

$$X_j = \sum_{j=1}^{N} g_j e_j + m \qquad (9)$$

Here, some properties of eigenvectors and eigenvalues derived from PCA are worth mentioning. First, all eigenvectors in the matrix $\phi$ are orthogonal, which indicates that the components calculated by PCA are not correlated with each other. Moreover, in general, once eigenvectors are found from the covariance matrix, the next step is to order them by eigenvalue, from highest to lowest. We know that the sum of the eigenvalues is equal to the total variance in the data. Therefore, the first component accounts the largest amount of the variance in the data. The ranking of eigenvalue shows the significance of the corresponding eigenvectors from which we can generate the criteria of components selection.

### 2.3.2. Algorithm design

In the application process of nonbinary hypointensity description (Figure 6), a matrix ROI is formed according to pixel values in the ROIs, and within the matrix each row vector stands for an individual image.

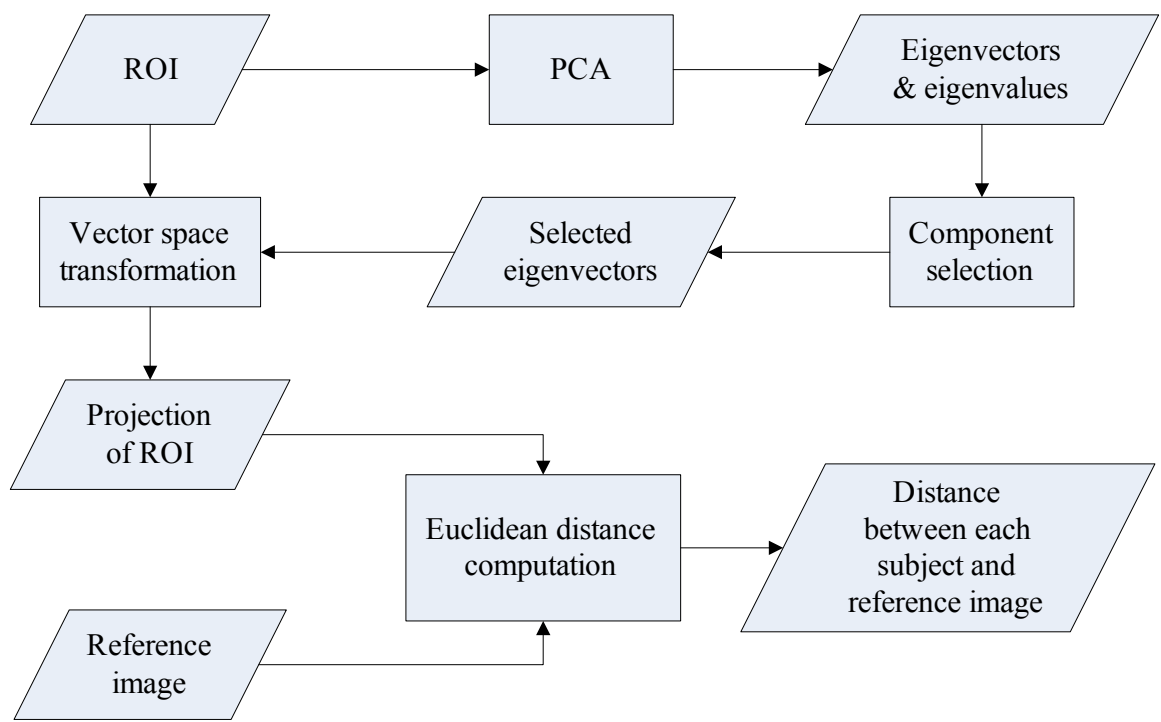

Figure 6: The application process of nonbinary hypointensity description

Then we use PCA to get eigenvectors and eigenvalues from the covariance of normalized ROI. In order to eliminate redundant information, a component selection process is needed. This number of selected eigenvectors is determined by a percentage of total variance we choose to keep in the projection of original data. After several tests with various percentage values we selected 70% as the proportion of variance to retain, because it gives the highest agreement when comparing its nonbinary description result

 

with the ground truth. Note that, the comparison approach will be introduced in the following section. Afterwards, the number of selected eigenvectors $n$ can be easily calculated according to Eqn. 10, where $N$ is the number of total eigenvectors extracted.

$$70\% = \frac{\sum_{i=1}^{n} eigenvalue_i}{\sum_{i=1}^{N} eigenvalue_i} \qquad (10)$$

Next, having both the eigenvectors and the ROI we get projections of ROI in the eigen space using Eqn. 8. At this time, original ROI which has $N$ dimensions has been transformed into the eigen space with less ($n$) dimensions while keeping 70% of the total variance. Then, we choose the image which has the highest HypoLoad according to the binary hypointensity description, and regard it as our reference image. Finally, Euclidean distance in eigen space between the reference image and other images are computed and used as our nonbinary hypointensity description.

This description is easy to comprehend, in binary hypointensity description the image with the highest HypoLoad is the one that has the darkest basal ganglia region. Consequently, image which has shorter distances to the reference image in nonbinary hypointensity description is more alike the reference image with respect to intensity that is relatively darker than the other images. Similarly, image with the lowest HypoLoad (brightest basal ganglia region) can be taken as the reference image as well, by calculating the distance in a reverse way.

Actually, we attempted to use Fisher Linear Discriminant Analysis (Fisher LDA)[17][18] as a comparison, which is a technique used to find the linear combination of features that best separate two or more classes of objects or events. It is closely related to PCA in that both look for linear combinations of variables which best explain the data. Fisher LDA explicitly attempts to model the difference between the classes of data, PCA on the other hand does not take into account any difference in classes. However, during the application procedure the results we gained by Fisher LDA were not as consistent as we expected. In case it takes too much time to verify the feasibility of this method on our dataset we shifted all our focus on PCA.

# 3.    Algorithm development

Previously, to localize the basal ganglia a Talairach brain atlas based approach was used, which tessellates the minimum bounding rectangle of the brain tissue in each slice into rectangular regions[2]. Accordingly, certain grid locations of the atlas indicated the position of basal ganglia structures. Our observations, however, showed that with this atlas-based approach the boundary regions introduced inconsistency to hypointensity detection, most probably due to the imperfect structure localization. Therefore, in this work we opted for manual delineations of the basal ganglia by an expert.

## 3.1.    Fixed ROI

At first stage, we employed fixed ROI which means the expert only delineates ROI boundary on one image and then this boundary is replicated on other images in the dataset, as shown in Figure 7. The blue line in Figure 7(a) is the boundary of basal ganglia region defined by the expert. Such boundary is applied on another two images in the dataset as shown in Figure 7(b) and (c). The advantage of this approach is that it is easy to implement, as the expert only has to define the boundary once. However, it is not very accurate due to the spatial registration artifacts, even if all susceptibility images were registered to each other. Sometimes the non-ROI part may be over segmented while the ROI part may be out of the boundary. In either situation we would obtain inaccurate input data, which will directly affect the hypointensity description result.

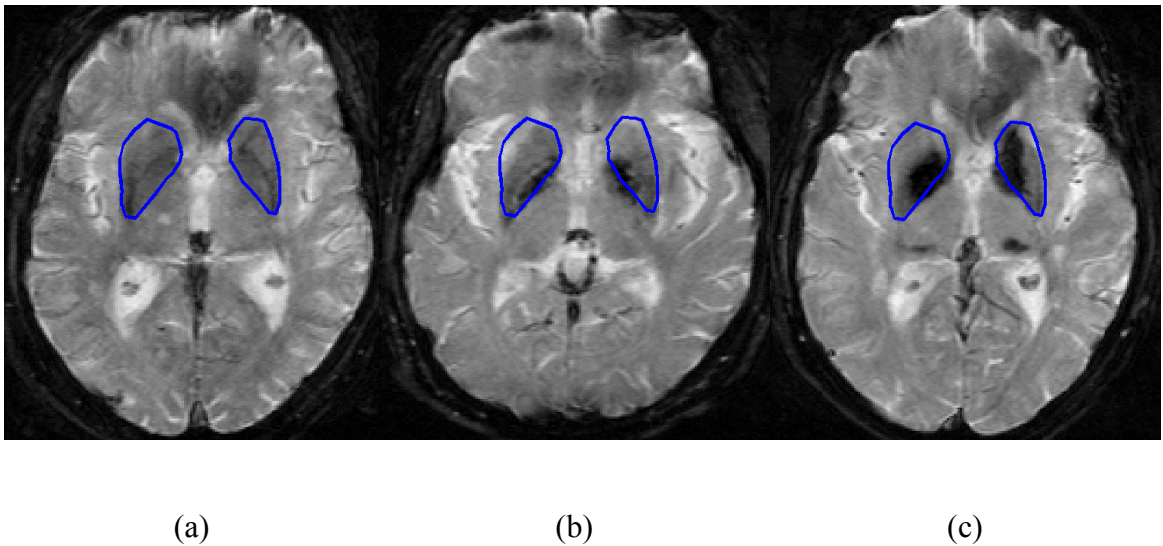

(a)                              (b)                              (c)

Figure 7: Examples of using fixed ROI

                                    

## 3.2. Personalized ROIs

In order to obtain more representative input data to achieve better performance of both binary and nonbinary hypointensity descriptions, we used personalized ROIs in the second stage. Personalized ROIs means all ROI boundaries are manually delineated by the expert to more perfectly fit the exact basal ganglia region. As illustrated in Figure 8, each image has its unique boundary which includes the basal ganglia region we are interested in without inaccurate segmentation. This method will directly increase the total accuracy, nevertheless, it is quite time consuming for the expert. Therefore, this method is only appropriate for small datasets such as in our case, if there are too many subjects in the dataset this method is not preferred.

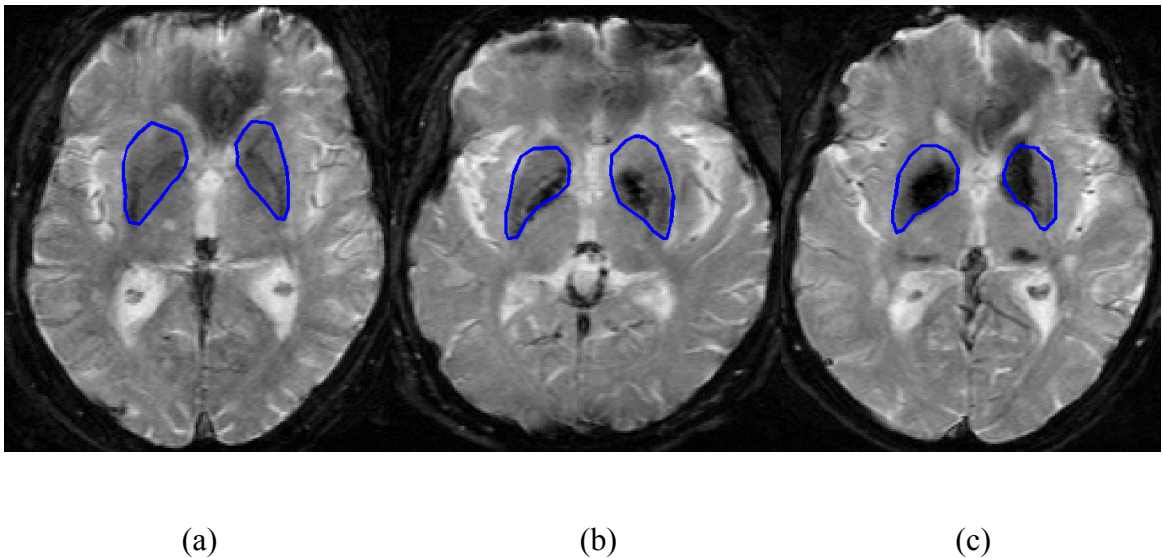

(a)                  (b)                  (c)

Figure 8: Examples of using personalized ROI

Although manual delineation of basal ganglia region will cost tremendous time, please note that, the proposed hypointensity description methods can be fully-automated by replacing this manual work by an automatic and accurate structure localization technique. In fact, there exists some recent preliminary work on shape-based automatic segmentation of basal ganglia structures[19]. We intend to employ such techniques in our framework in the future.

Another thing worth mentioning is that, as sizes of ROIs vary due to personalized delineation, we need to sample same number of pixels in all ROIs to form the ROI matrix used in PCA algorithm. This number is determined by the size of minimal ROI, and it can be simply obtained by an iteration of all ROIs. Once this number is given, sampling is needed to select certain number of pixels from every ROI. In the following we explain the sampling method used.

1.  Traverse an ROI in sequential raster manner and put every pixel in a row vector

*e.g. generated row vector:*

| 1 | 2 | 3 | 4 | 5 | 6 | 7 | 8 | 9 | 10 | 11 | 12 |
|---|---|---|---|---|---|---|---|---|----|----|----|

2.  Make all pixels in the row vector disordered

*e.g. disordered row vector:*

| 8 | 2 | 1 | 6 | 7 | 12 | 10 | 11 | 9 | 5 | 3 | 4 |
|---|---|---|---|---|----|----|----|---|---|---|---|

3.  Select as many pixels in the front of the row vector as in the minimal ROI

*e.g. selected pixels, and the minimal ROI has 8 pixels:*

| 8 | 2 | 1 | 6 | 7 | 12 | 10 | 11 |
|---|---|---|---|---|----|----|----|

4.  Sort selected pixels in descending order for the purpose of applying PCA on it

*e.g. sort selected pixels:*

| 1 | 2 | 6 | 7 | 8 | 10 | 11 | 12 |
|---|---|---|---|---|----|----|----|

The disadvantage of this sampling method is that it could cause inhomogeneity issue, which means the selected pixels may not evenly distributed (spatially unbalanced) in original ROI. Hence, we need to find a better solution to ensure the representativeness of sampled pixels.

## 3.3. Optimized description

### 3.3.1. Adaptive threshold

As discussed above, the reference region thresholding approach is not reliable enough. Accordingly, we proposed the adaptive thresholding method to improve the overall accuracy of binary hypointensity description.

First we normalized the ROIs by subtracting the mean from intensity data *ROI* and then dividing the result by the mean, as shown in Eqn. 11.

 

$$ROI_{-norm} = \frac{ROI - mean(ROI)}{mean(ROI)} \qquad (11)$$

From this normalization, we obtain a variation range of normalized intensity values as well as the distribution of the values. In the next step, an adaptive threshold is generated for hypointensity quantification, as illustrated in Figure 9.

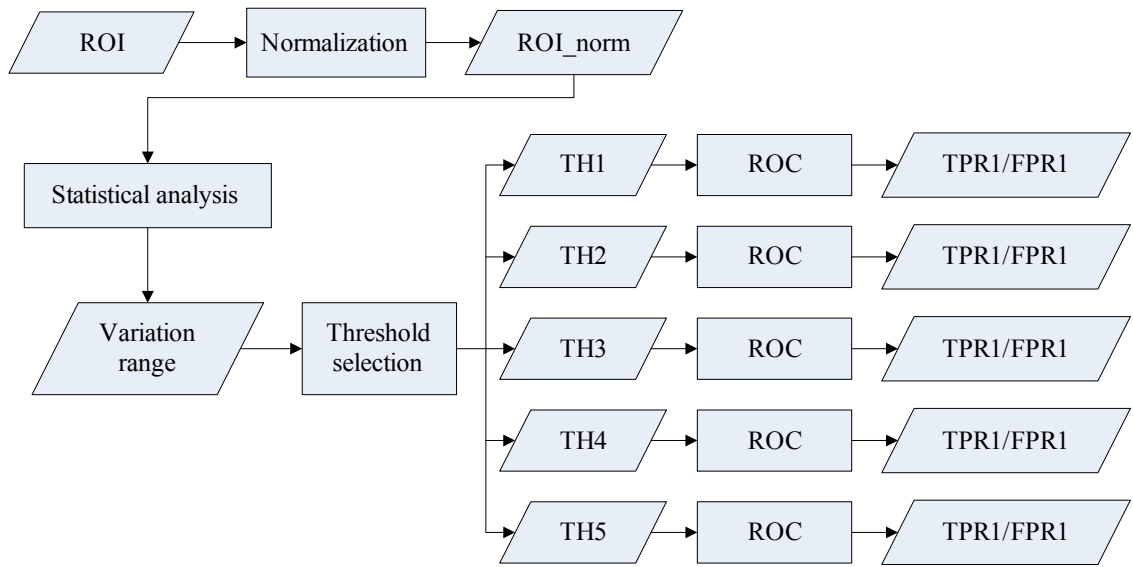

Figure 9: Application process of adaptive threshold

First we select several values in this variation range, considering them as thresholds and apply them to for each MR image. In our application, pixels which have values less than the threshold are regarded as hypointense (darker), and hypointensity is defined as percentage of hypointense pixels in ROI. In this way, each threshold gives one hypointensity result for each MR image, and then we sort these results in descending order to get a ranking. Next, we use this ranking to categorize all subjects into three clusters, the same as the ground truth. In Section 5, details on this clustering will be introduced, and also the way to compare the clustering results and the ground truth. Given these two clustering results, fractions of true positives (TPR) and false positives (FPR) are computed to estimate the agreement between each ranking and expert's clustering. In the end, the threshold which has the highest TPR and the lowest FPR is chosen and the hypointensities defined by this threshold are assigned as the final binary hypointensity description. In the evaluation part, it will be convinced that this way of defining threshold is more reasonable and robust.

Note that, when a new subject arises, different solutions should be applied to different situations. If the threshold is generated from a small dataset, such as in our case, re-calculation of adaptive threshold is necessary, as the newly input data probably will change the TPR and FPR of each candidate threshold. But if the representativeness of the dataset is convinced, then there is no need to compute the threshold again. Furthermore, this whole procedure is automated.

### 3.3.2. Spatially balanced sampling

As explained in 3.2, pixel sampling is needed so that PCA can be applied on personalized ROIs. However, in previous sampling method we did not consider the distribution of selected pixels, thus it has a drawback of inhomogeneity. And in our application the representativeness of inputted ROI pixels is critical, because selection of either too many hypointense pixels or non-hypointensity pixels will affect the accuracy of the final result. Therefore, we propose the spatially balanced sampling to overcome this disadvantage and obtain more representative ROI pixels. Following introduces the idea of this sampling method via an example.

In spatially balanced sampling, the first and last pixel of current ROI are kept while the rest of the pixels are generated using interpolation, as illustrated in Figure 10.

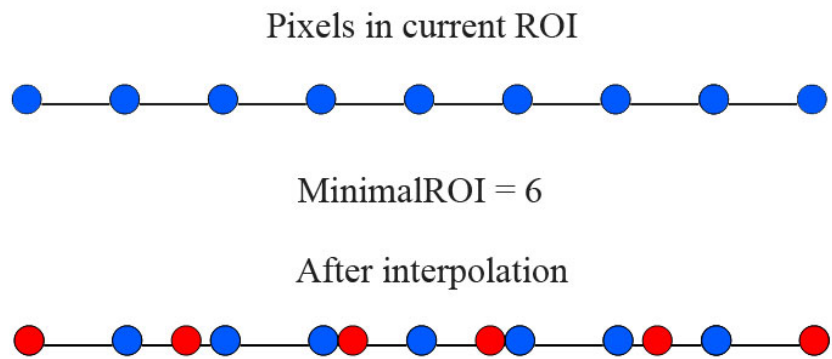

Figure 10: Example of spatially balanced sampling

In Figure 10, blue ball represents pixels in current ROI and red ball represents pixels of current ROI after sampling. In the example, current ROI has 9 pixels while the size of minimal ROI is 6. In order to make the sampled pixels (red balls) evenly distributed, besides the first and last sampled pixels (red balls), other sampled pixels (red balls) are calculated according to their neighboring original pixels (blue balls). And the distance between every nerghboring sampled pixels (red balls) is computed using Eqn. 12.

$$stepsize = \frac{N_{currentROI} - 1}{N_{MinROI} - 1}$$ ( 12 )

Once the position of one sampled pixel (red ball) with respect to its neighboring original pixels (blue balls) is known, the pixel value of this pixel (red ball) can be calculated in a way that the nearer original pixel (blue ball) value gives more proportion to the final resu lt.

 

# 4. Feature correlation analysis

## 4.1. Feature definition

In a previous study based on binary description of hypointensity in the brain, it was shown that brain iron accumulation shape provides additional information to the shape-insensitive features[2], such as the total brain iron load, that are commonly used in clinics. Therefore, in this section we define both binary and nonbinary features and compare the complementary and redundant information provided by the two descriptions using Kendall's rank correlation coefficient.

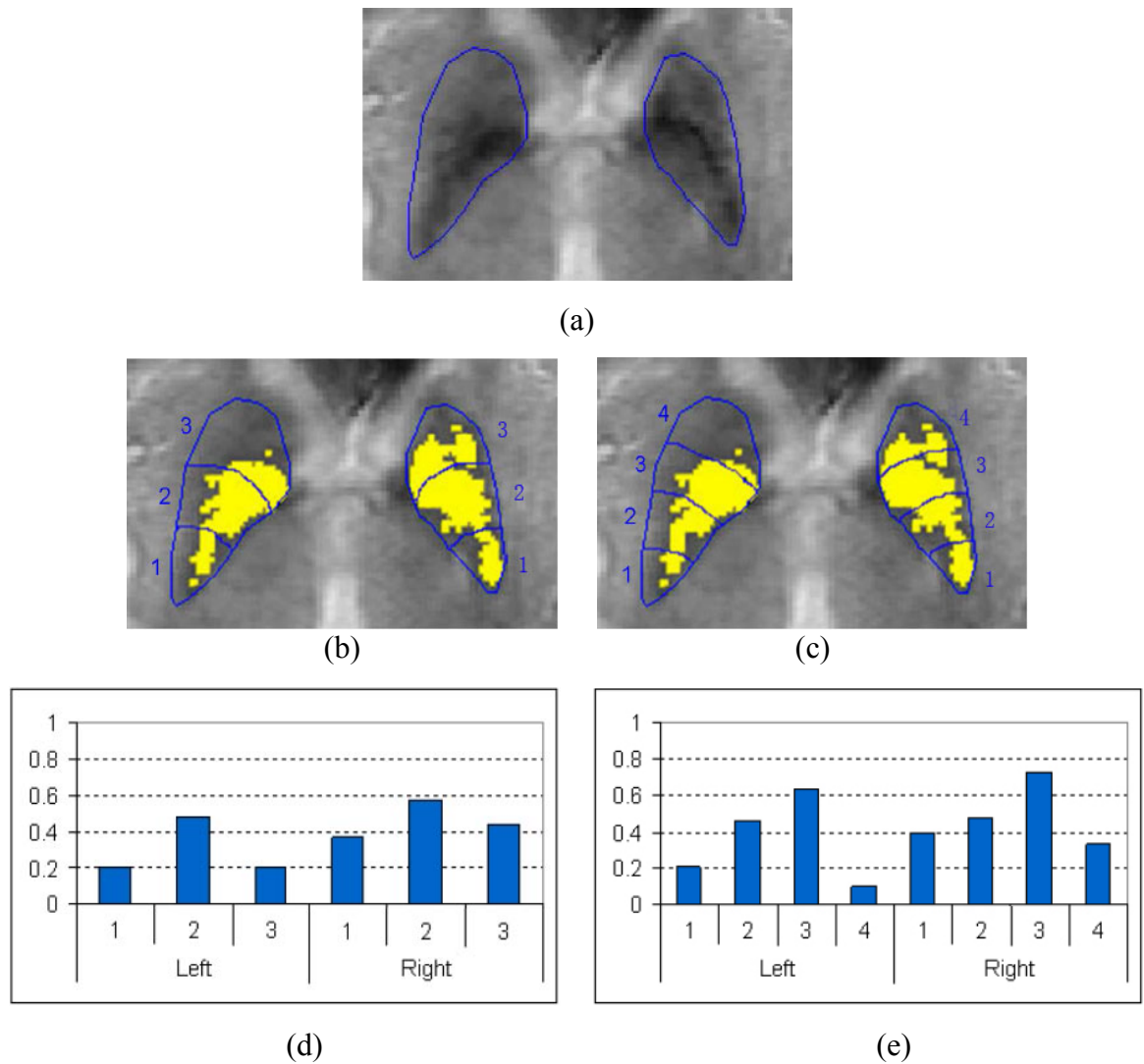

Figure 11: Binary hypointensity feature example. (a): Zoomed SWI image of one basal ganglia delineation. (b)(c): Same basal ganglia tessellated into radially equidistant subregions (left: N=3, right: N=4) and hypointense pixels highlighted. (e)(f): Corresponding subregion hypointensities displayed in graphs.

For the binary hypointensity feature, in order to capture the spatial distribution of iron in basal ganglia we tessellate the ROIs as shown in Figure 11. Each hemispheric part of the ROI is divided into $N$ radially equidistant subregions with the outermost posterior pixel selected as the central point. Please note that, we define radially equidistant subregions by concentric circles with $(r_{i+1} - r_i) = \Delta r$, where $\Delta r$ is constant and $i$ is a positive integer. Exemplary tessellations for $N = 3$ and $N = 4$, and the corresponding subregion hypointensities are displayed in Figure 11. And the binary feature is the percentages of hypointense pixels in each basal ganglia subregion.

For nonbinary hypointensity feature, our initial observations showed that images of basal ganglia with varying hypointensities have distinctive eigenspace projections, as shown in Figure 12. Therefore, we define nonbinary hypointensity features of an ROI as the corresponding eigenspace projection values from each principal component.

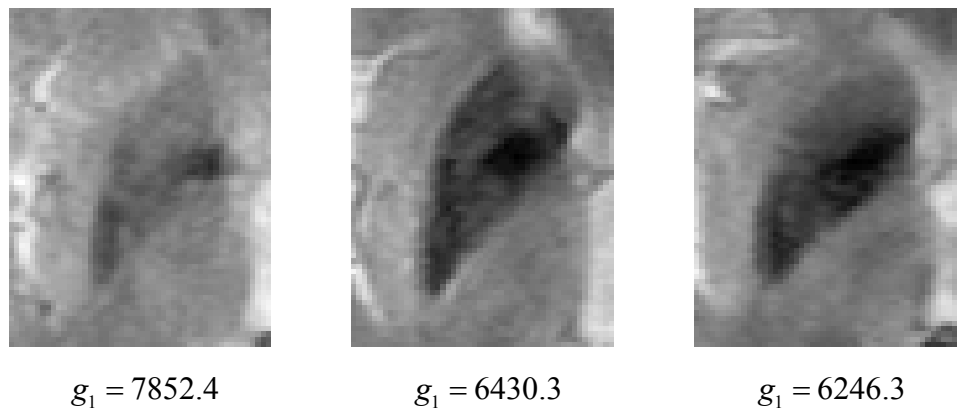

$g_1 = 7852.4$ $\qquad$ $g_1 = 6430.3$ $\qquad$ $g_1 = 6246.3$

Figure 12: Exemplary SWI images with varying hypointensities in the basal ganglia and the corresponding eigenspace projection value, $g$, from the first principle component.

## 4.2. Kendall's correlation coeffient

For the purpose of exploring the correlation among the proposed binary and nonbinary hypointensity features in the previous part. We propose to use Kendall's rank correlation coefficient, also referred to as Kendall's tau, that measures the correlation between two rankings as a non-parametric statistical measure[20]. This is a very convenient way to compare the performance of search and retrieval methodologies because the returns to various queries in the domain of search and retrieval are often represented as the (similarity) ranks of the items in the database. Higher values of Kendall's tau show that the two ranks are correlated; hence, in our case, those features that have higher correlation value between each other possess similar information.

Kendall's tau between feature A and B is computed as follows:

*For each subject in the dataset used as query*

 

1. *Select on of the features as feature A and another one as feature B*

2. *Find the ranked list (Rank A) of the returns to the query with feature A*

3. *Find the ranked list (Rank B) of the returns to the query with feature B*

4. *Match Rank B with Rank A to find the respective position of same dataset items. For example,*

| Subjects | a | b | c | d | e | f | g | h |
|---|---|---|---|---|---|---|---|---|
| Rank by Feature A | 1 | 2 | 3 | 4 | 5 | 6 | 7 | 8 |
| Rank by Feature B | 3 | 4 | 1 | 2 | 5 | 7 | 8 | 6 |

5. *For each subject in the Rank B row, compute the number of entries on its right that have higher rank and sum these numbers to find the P value*

6. *For each subject in the Rank B row, compute the number of entries on its right that have lower rank and sum these numbers to find the Q value*

7. *Having N as the total number of entries (eight in the above example), compute tau as:*

$$tau = \frac{P - Q}{\frac{1}{2} \times N \times (N-1)}$$

( 13 )

Finally, for each feature pair we repeat the above steps and assign the average of the individual values computed as the corresponding Kendall's correlation value.

## 4.3. Analysis results

### 4.3.1. Single tessellation

In order to analyze the relationships between individual features we first introduce the results of a single tessellation[21]. Figure 13 visualizes the resulting correlation measurements for the binary, nonbinary and binary versus nonbinary features that are computed from the tessellation size of 10 using a heat map – a data visualization technique based on color. Regarding the binary features, we observe medium-strength (in the range of [0.4-0.7]) correlation between neighboring features (or subregions), and this correlation weakens as the proximity between features increases. Nonbinary feature results show weak (below 0.4) correlations between eigenvectors, which is consistent with the fact that PCA finds a transformation that uncorrelates the variables. When we compare binary features with the nonbinary ones we see that the first nonbinary feature, which is the eigenvector with the largest variance, exhibits medium correlation with binary features, while the rest have weak figures.

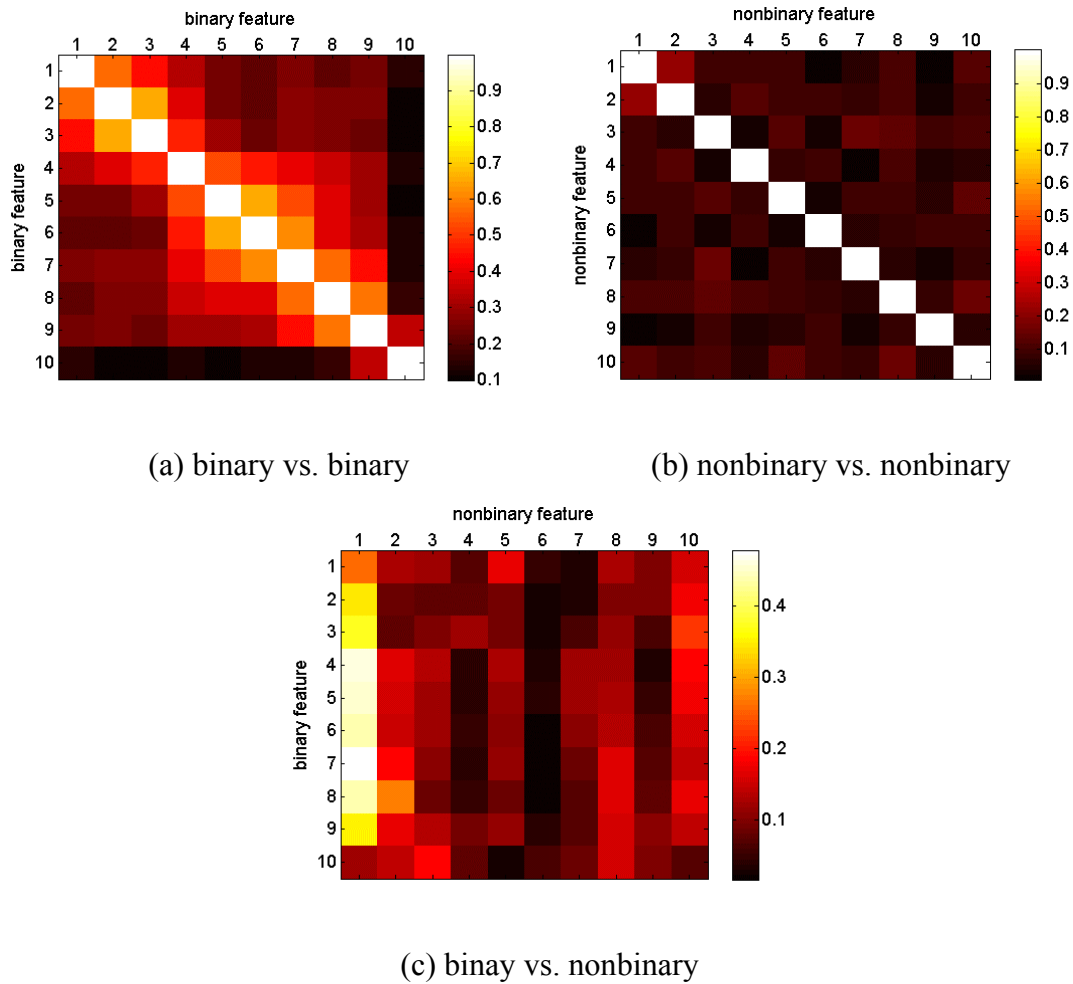

(a) binary vs. binary   (b) nonbinary vs. nonbinary

(c) binay vs. nonbinary

Figure 13: Heat maps of Kendall's correlation values for single tessellation.

### 4.3.2.  Multiple tessellations

"Does the tessellation size affect correlations?" To answer this question, in this subsection we present the results of multiple tessellations[21]. Contrary to the previous subsection, here we use combination of features to describe hypointensity and refer to them as descriptions (e.g. binary description=4 refers to a 4-dimensional vector of hypointensities computed from 4 basal ganglia subregions, or nonbinary description=7 means 7 eigenvectors with highest variance are kept).

Figure 14 visualizes the corresponding averaged correlation values over multiple runs for the binary-only, nonbinary-only and binary versus nonbinary descriptions, respectively. The analysis reveals that within-description (binary-only and nonbinary-only) correlations are strong (above 0.7), while between-description (binary versus nonbianry) correlations have medium strength. Furthermore, in the latter, we observe that tessellation size has no considerable effect on the results, meaning spatial and eigenspace

   

tessellations interrelate with each other. We conclude that combination of the two descriptions may provide a better representation of hypointensity. Hence further evaluation is needed to find the optimal tessellation and combination of descriptions.

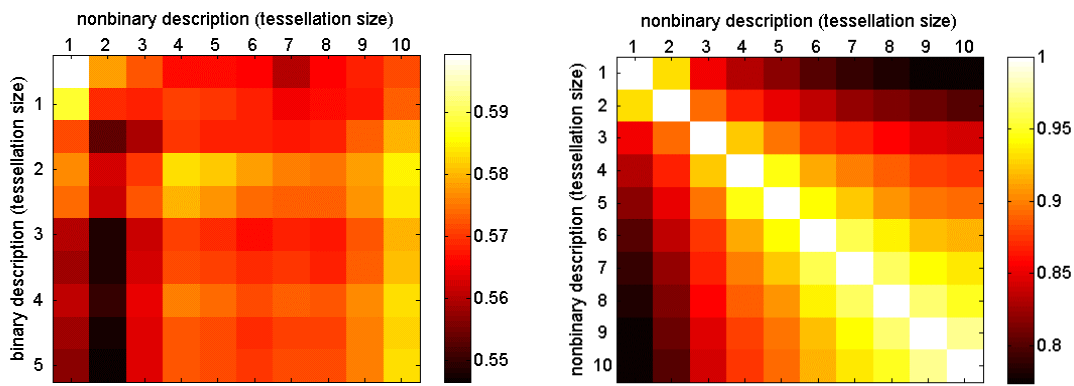

(a) binary vs. binary        (b) nonbinary vs. nonbinary

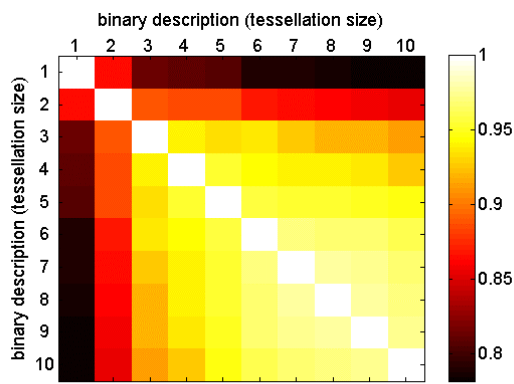

(c) binary vs. nonbinary

Figure 14: Heat maps of Kendall's correlation values for multiple tessellations.

# 5.    Evaluation

## 5.1.    Evaluation methodology

In order to evaluate the performances of the proposed hypointensity descriptions, we compared their ranking results with the ground truth. To do this, every ranking result is also divided into three clusters, the ones at the top of the ranking are categorized as light while the ones at the bottom as dark. For each cluster the number of included subjects is the same as that of the ground truth, as illustrated in Table 1.

| Cluster | Ground truth | Ranking to be evaluated | # of in common |
|---------|--------------|-------------------------|----------------|
| Light | 2 | 2 | |
| | 7 | 4 | |
| | 8 | 5 | |
| | 10 | 6 | |
| | 14 | 8 | |
| | 16 | 10 | |
| | 17 | 13 | 7 |
| | 18 | 14 | |
| | 21 | 17 | |
| | 22 | 20 | |
| | 24 | 21 | |
| | 33 | 27 | |
| | 26 | 33 | |
| Mid | 1 | 1 | |
| | 3 | 3 | |
| | 4 | 7 | |
| | 6 | 11 | |
| | 11 | 12 | |
| | 12 | 18 | |
| | 13 | 22 | 6 |
| | 20 | 23 | |
| | 23 | 24 | |
| | 27 | 25 | |
| | 29 | 26 | |
| | 32 | 30 | |
| | 35 | 32 | |
| Dark | 5 | 9 | |
| | 9 | 15 | |
| | 15 | 16 | |
| | 19 | 19 | |
| | 25 | 28 | |
| | 26 | 29 | 6 |
| | 28 | 31 | |
| | 30 | 34 | |
| | 31 | 35 | |
| | 34 | 36 | |
| | 37 | 37 | |

Table 1: Comparing the ranking result with the ground truth. Common items are highlighted with color blue, yellow and green for dark, mid and light clusters respectively.

    

The second column shows the index of subjects in each cluster, and the third column is a ranking to be evaluated. Then we count the number of identical subjects in every cluster, as highlighted with different colors in Table 1. Given these numbers, the accuracy of the evaluated ranking can be calculated using Eqn. 14.

$$Accuracy = \frac{\dfrac{n_l}{N_l} + \dfrac{n_m}{N_m} + \dfrac{n_d}{N_d}}{3} \qquad (\,14\,)$$

Where $n_l$ , $n_m$ , $n_d$ are the number of common subjects in clusters light, mid, dark respectively, and $N_l$ , $N_m$ , $N_d$ are the total number of subjects contained in each cluster.

## 5.2.  Test results

Based on the evaluation method explained above, we calculated the agreement between the ground truth and each clustering result. These results are generated by using different versions of both binary and nonbinary hypointensity descriptions, referred to as *HypoLoad* and *PCA* respectively. Moreover, for the purpose of testing the consitency of the proposed methods, we applied them on each hemisphere separately and on the whole brain.

Table 2 shows the agreements of both *HypoLoad* and *PCA* with the ground truth when using fixed ROI, and the average accuracy of each method is displayed in the last row (Accuracy). The comparison in left, right hemisphere and the whole brain illustrates that *PCA* outperforms *HypoLoad* with approximately 20% higher accuracy.

| # of in common subjects | *Left* | | *Right* | | *Whole* | |
| | *HypoLoad* | *PCA* | *HypoLoad* | *PCA* | *HypoLoad* | *PCA* |
|---|---|---|---|---|---|---|
| Light | 4  31% | 10  77% | 5  38% | 10  77% | 5  38% | 10  77% |
| Mid | 5  38% | 6  46% | 2  15% | 8  62% | 5  38% | 8  62% |
| Dark | 8  73% | 7  64% | 7  64% | 9  82% | 8  73% | 9  82% |
| Accuracy | **45%** | **64%** | **38%** | **74%** | **48%** | **74%** |

Table 2: Fixed ROI results.

  

Table 3 shows the agreements of both *HypoLoad* and *PCA* with the ground truth by using personalized ROI with average accuracy results of fixed ROI (Accuracy') displayed in the last row.

| # of in common subjects | *Left* | | *Right* | | *Whole* | |
|---|---|---|---|---|---|---|
| | *HypoLoad* | *PCA* | *HypoLoad* | *PCA* | *HypoLoad* | *PCA* |
| Light | 7    54% | 9    69% | 8    62% | 9    69% | 8    62% | 9    69% |
| Mid | 6    46% | 9    69% | 3    23% | 10    77% | 5    38% | 9    69% |
| Dark | 7    64% | 11    100% | 6    55% | 11    100% | 7    64% | 11    100% |
| **Accuracy** | **55%** | **79%** | **47%** | **81%** | **55%** | **79%** |
| Accuracy' | 45% | 64% | 38% | 74% | 48% | 74% |

Table 3: Personalized ROIs results.

From Table 3, it is straightforward to conclude that personalized ROIs gives more accurate input data which directly affects the results of both *HypoLoad* and *PCA*. By manually delineating boundaries of ROI for each subject, the accuracy of *HypoLoad* improves about 8% and *PCA* improves at least 5%. Even though *PCA* performs better than *HypoLoad* in all considered cases.

The agreements of both *HypoLoad* and *PCA* with the ground truth when using optimized description, namely employs adaptive threshold in *HypoLoad* and spatially balanced sampling in *PCA*, is shown in Table 4. Same as in Table 3, the average accuracy of previous version of methods is exhibited in the last row.

| # of in common subjects | *Left* | | *Right* | | *Whole* | |
|---|---|---|---|---|---|---|
| | *HypoLoad* | *PCA* | *HypoLoad* | *PCA* | *HypoLoad* | *PCA* |
| Light | 9    69% | 10    77% | 9    69% | 10    77% | 10    77% | 10    77% |
| Mid | 9    69% | 10    77% | 8    62% | 10    77% | 10    77% | 10    77% |
| Dark | 11    100% | 11    100% | 10    91% | 11    100% | 11    100% | 11    100% |
| **Accuracy** | **79%** | **84%** | **73%** | **84%** | **84%** | **84%** |
| Accuracy' | 55% | 79% | 47% | 81% | 55% | 79% |

Table 4: Optimized description results.

Table 4 illustrates that by using adaptive threshold in *HypoLoad* method, the average accuracy has increased dramatically (roughly 20%), while spatially balanced sampling gives *PCA* about 3% improvement. Although when applying both methods on the whole brain, they achieved the same accuracy, as high as 84%, considering the overall performance of *PCA*, we can conclude that nonbinary hypointensity description outperforms binary hypointensity description in most cases.

                    © XIAOJING CHEN. 2008

As indicated in the above tables, our proposed nonbinary hypointensity description method *PCA* performs better than the binary method *HypoLoad*. In general, these results confirm the improvement of the optimized methods compared with the original ones, thus the validity of the developed methods, from an objective point of view, is verified.

In order to give a better illustration, the graph representation of the performances is shown in the following figure. The bars represent average accuracy, while the red vertical lines illustrate the span of accuracy in corresponding approaches and the numbers below are lengths of the spans.

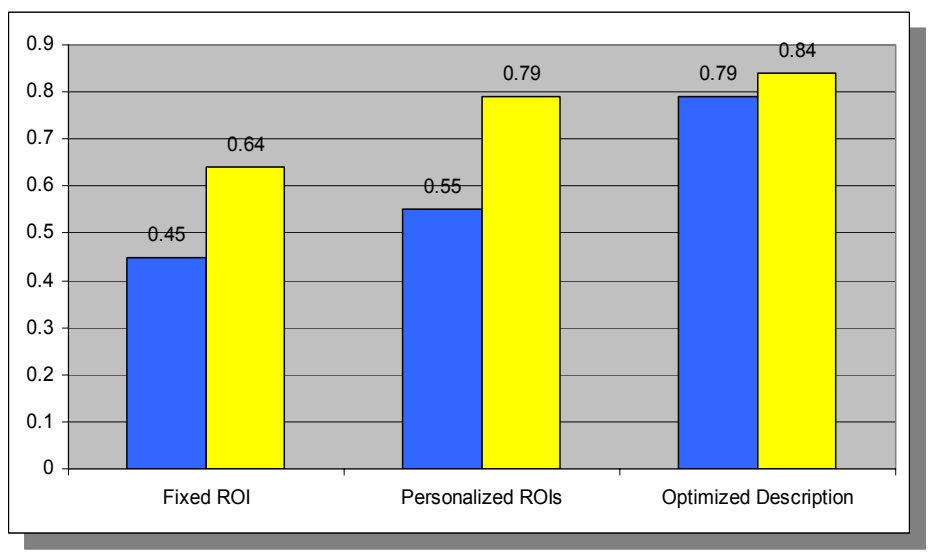

(a)

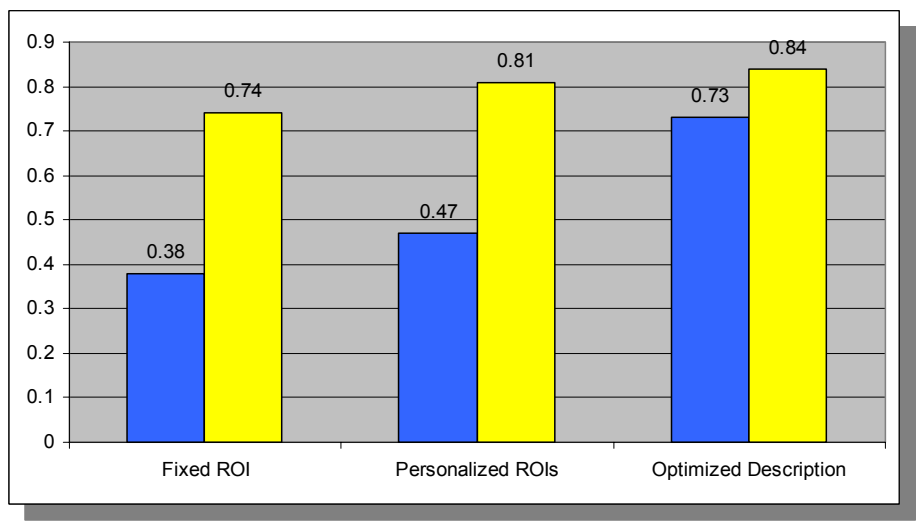

(b)

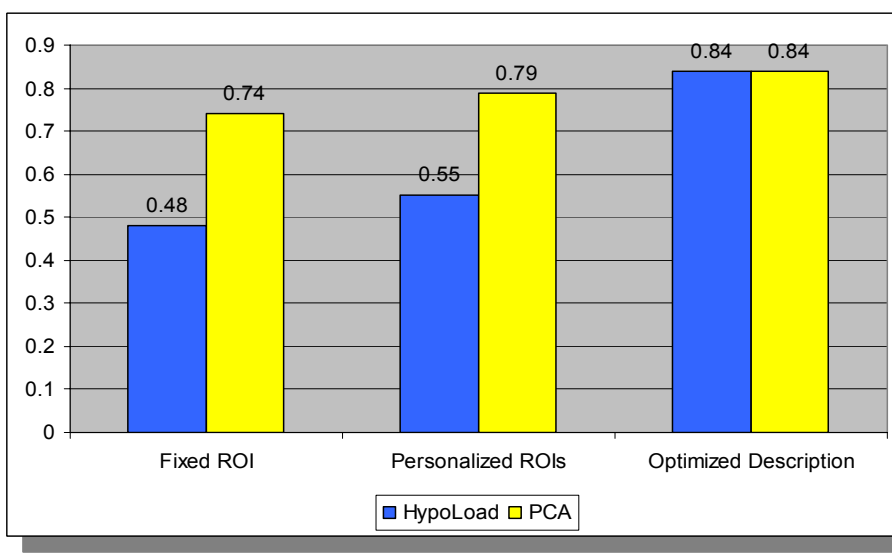

(c)

Figure 15: Graph representation of evaluation results. (a) Different versions of hypointensity methods applied on the left hemisphere, (b) hypointensity methods applied on the right hemisphere, (c) hypointensity methods applied on the whole brain.

Although optimized descriptions achieved the highest accuracy, as we explained before, in this version manual delineation of the basal ganglia region from an expert is required, which would significantly increase the whole time complexity. In our application, the required time to process (intensity variation correction and the ROI boundary delineation) one image is about 1 minute. If the expert were to manually delineate a VOI instead of a single slice and hundreds of subjects were in the dataset, then the required time would be much higher. Therefore, if a slightly lower accuracy (74% for the whole brain) is acceptable, *PCA* applied on fixed ROI is preferred. In order to determine the most suitable solution, a trade-off between time complexity and accuracy is inevitable.

## 5.3.    Evaluation with Golden data

Besides susceptibility-weighted images, we also applied our methods on Proton Density (PD)-T2 images, which were called Golden data in our project since the dataset included manual annotations of hypointensity at pixel-level made by a medical expert. In total, there are 20 subjects in the Golden dataset, and they were captured at LUMC (Leiden University Medical Center).

For each subject, we extracted one PD image from the same location of the brain scan as well as its corresponding T2 image and manual annotation image (Figure 16). Because PD image gave better visible boundary of the basal ganglia region, it was used for manually delineation of ROI. Both the binary and nonbinary hypointensity description methods were tested on T2 image, due to its sensitivity to iron. However, it is obvious that the intensity contrast of T2 is comparatively lower than that of susceptibility image shown in Figure 2(a). Moreover, the hypointensity pixels were marked in white for the whole brain by a medical expert in manual annotation image.

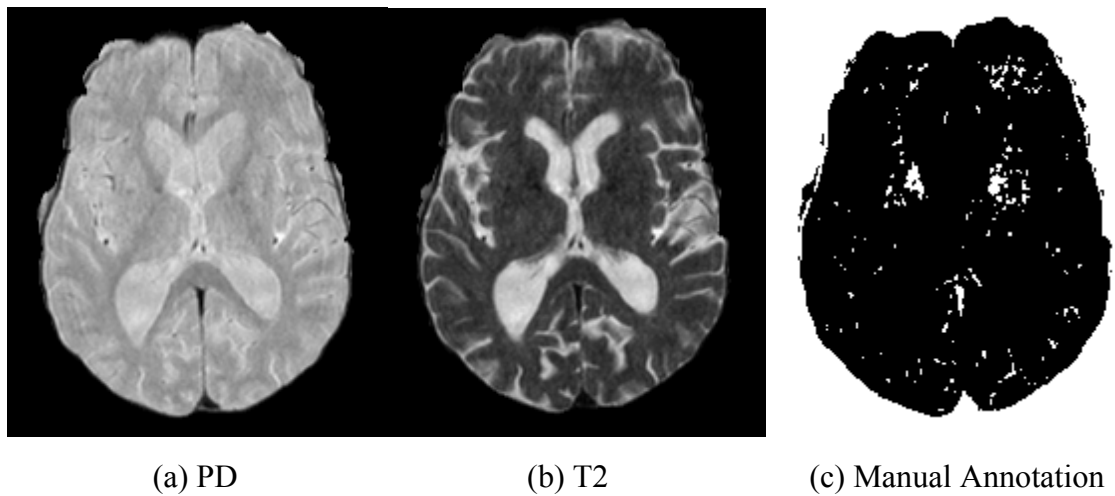

(a) PD                              (b) T2                         (c) Manual Annotation

Figure 16: Exemplary Golden data images.

Once all outlines of ROIs were obtained by manual delineation of the basal ganglia region on PD image, we replicated them on corresponding T2 and manual annotation images. From the manual annotation image the ratio of hypointensity pixels in the ROI was calculated by dividing the number of hypointense pixels by the total number of pixels in the ROI. All hypointensity ratios were put in descending order as illustrated in Figure 17. From this graph a sharp increase of hypointensity was observed between subject 13 and 14, thus it was employed as a threshold to separate all subjects into two

                                                              

clusters: light and dark. Accordingly, there were 13 subjects in cluster light and 7 subjects in cluster dark. We considered this clustering as the ground truth for Golden data. As annotation of hypointense pixels was accomplished by an expert at pixel-level, the ground truth for Golden data is more objective than the previous one. However, as confidence of this ground truth is much relied on the expert's notation accuracy, it may gave rise to potential errors in the ground truth.

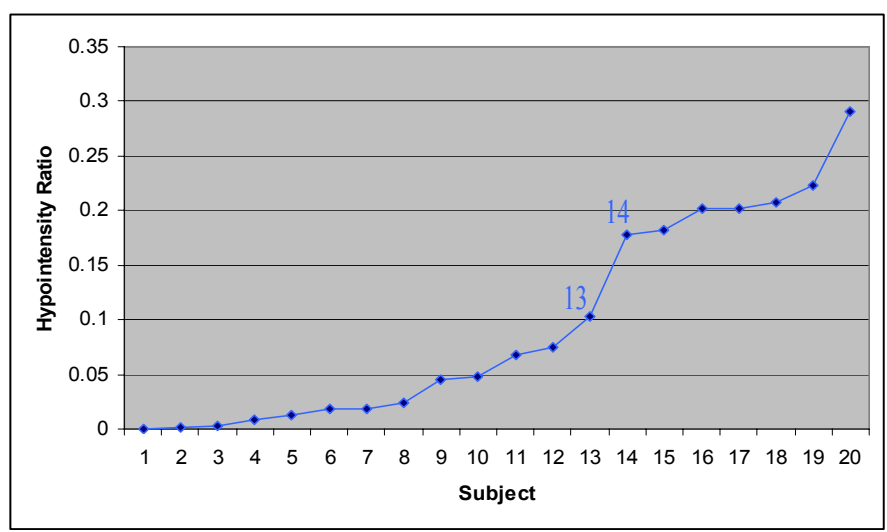

Figure 17: Hypointensities of all subjects.

After applying the optimized hypointensity description methods on each hemisphere and the whole brain separately, we compared the results with the Golden data ground truth. The agreements and average accuracy of both *HypoLoad* and *PCA* were obtained, as shown in Table 5.

| # of in common subjects | Left | | Right | | Whole | |
|---|---|---|---|---|---|---|
| | *HypoLoad* | *PCA* | *HypoLoad* | *PCA* | *HypoLoad* | *PCA* |
| Light | 11  85% | 11  85% | 11  85% | 10  77% | 11  85% | 11  85% |
| Dark | 5  71% | 5  71% | 5  71% | 4  57% | 5  71% | 5  71% |
| Accuracy | **78%** | **78%** | **78%** | **67%** | **78%** | **78%** |

Table 5: Golden data results.

The data revealed that *HypoLoad* and *PCA* achieved approximately the same accuracy, around 78%. However, the performance of *PCA* on T2 images was not as good as that on susceptibility-weighted images, which is probably due to the lower intensity contrast in T2 images. Nevertheless, the results confirmed the reliability and robustness of both hypointensity description methods on different MR contrasts.

# 6.    Conclusions and Recommendations

## 6.1.    Conclusions

Although iron accumulation in human brain is a normal process that starts at the early ages and is perceived in every individual, more and more evidence has shown that in those developing neurodegenerative diseases abnormal (larger) deposition of iron appears, especially in the basal ganglia region. Due to the changes in MR signal caused by presence of iron, tissues with high iron concentration appear hypo-intense (darker than usual) in MR contrasts, such as T2 and T2*, which are traditional MR contrasts used by medical experts to visualize brain and differentiate various tissues. Unlike these traditional contrasts which focus mainly on the signal magnitude, susceptibility-weighted imaging (SWI), a novel MR contrast, exploits both magnitude and phase information. More importantly, it has higher sensitivity to magnetic materials, such as iron.

Therefore in this work, we proposed reliable and robust methodologies to describe hypointensity in the basal ganglia region of human brain. We developed a binary description approach by replacing the Talairach brain atlas with manual delineation to more accurately localize the region-of-interests (ROIs). Furthermore, a novel nonbinary hypointensity description method based on principle component analysis was constructed. Gradual improvement of the algorithms was carried out not only by obtaining more representative input data but also via optimizing the algorithms themselves.

Besides that, several hypointensity features were defined in both binary and nonbinary descriptions. We compared the complementary and redundant information provided by these features using Kendall's rank correlation coefficient in order to better understand the individual descriptions of iron accumulation. The analysis revealed high within-feature (binary-only and nonbinary-only) correlations, while the between-feature (binary versus nonbinary) correlations were at medium strength. Moreover, we observed that the number of features does not have considerable impact on the results, which confirmed that binary (shape-related) and nonbinary (eigenspace-related) descriptions correlate with each other.

Through comparison between the results of different versions of both descriptions and the ground truth of SWI dataset, nonbinary approach based on PCA when using spatially balanced sampling method achieved the highest accuracy. Nonbinary approach using fixed ROI would be preferred for the cases where lower accuracy is acceptable and less time complexity is required. Performance evaluation of both descriptions was carried out on a Golden dataset as well. The results further confirmed the reliability and robustness of our proposed methods.

                                    

## 6.2. Recommendations

In this report, the ground truth we used for each dataset to evaluate the performaces of hypointensity descriptions involved different degrees of subjectiveness. Because each ground truth was created according to the medical experts' judgement, which may include subjective errors. Hence, we will consider to use more objective information as our ground truth in the future.

The possible future work may further involve the investigation of the optimum ROI localization. Currently, the boundary of ROI is manually delineated by the expert which costs tremendous time and contains segmentation inaccuracies. With a more advanced segmentation technique of human brain structures, our proposed approaches could achieve higher accuracy and can be realized in a fully automated fashion.

Moreover, improved evaluation of the algorithms will also be helpful in providing more convincing results to prove the reliability and robustness of proposed methods. Finally, we would like to extend the application of both binary and nonbinary hypointensity descriptions on 3D data in the future.

# References

[1] "2008 Alzheimer's disease fact and figures", Alzheimer's Association, pp. 9-13, 2008.

[2] Ekin A, Jasinschi R, Turan E, Engbers R, Grond J, and Buchem M, "Search and retrieval of medical images for improved diagnosis of neurodegenerative diseases", Proc. SPIE Electronic Imaging, Volume 6506, 650604, 2007

[3] Hallgren B, Sourander P, "The effect of age on the non-haemin iron in the human brain", J Neurochem 3, pp. 41-51, 1958

[4] Loeffler DA, Conner JR, Juneau PL, Snyder BS, Kanaley L, DeMaggio AJ, et al., "Transferrin and iron in normal, Alzheimer's disease, and Parkinson's diseas brain regions", J. Neurochem 65, pp. 710-724, 1995

[5] Bartzokis G, Beckson M, Hance DB, Marx P, Foster JA, Marder SR, "MR evaluation of age-related increase of brain iron in young adult and older normal males", Magn. Reson. Imaging 15, pp. 29-35, 1997

[6] Marin WR, Ye FQ, Allen PS, "Increasing striatal iron content associated with normal aging", Mov. Disord. 13, pp. 281-286, 1998

[7] Ogg RJ, Langston JW, Haacke EM, Steen RG, Taylor JS, "The correlation between phase shifts in gradient-echo MR imaging and regional brain iron concentration", Magn. Reson. Imaging 17, pp. 1141-1148, 1999

[8] Harder SL, Hopp KM, Ward H, Neglio H, Gitlin J, and Kido D, "Mineralization of the deep gray matter with age: a retrospective review with susceptibility-weighted MR imaging", American Journal of Neuroradiology 29, pp. 176-183, 2008

[9] Thomas M, Jankovic J, "Neurodegenerative disease and iron storage in the brain", Curr. Opin. Neurol. 17, pp. 437-442, 2004

[10] Zecca L, Youdim MB, Riederer P, Connor JR, Crichton RR, "Iron, brain ageing and neurodegenerative disorders", Nat. Rev. Neurosci. 5, pp. 863-873, 2004

[11] Lauterbur PC, "Image formation by induced local interactions: examples of employing nuclear magnetic resonance". Nature 242, pp. 190–191, 1973

[12] Haacke EM, Xu Y, Cheng YCN, and Reichenbach JR, "Susceptibility weighted imaging (SWI)", Magnetic Resonance in Medicine 52, pp. 612-618, 2004

[13] Thomas B, Somasundaram S, Thamburaj K, Kesavadas C, Gupta AK, Bodhey NK, and Kapilamoorthy TR, "Clinical applications of susceptibility weighted MR imaging of the brain – a pictorial review", Diagnostic Neuroradiology 50, pp. 50:105-116, 2008

[14] Kostelec PJ and Periaswamy S, "Image registration for MRI", Modern Signal Processing, MSRI Publications, Volume 46, pp. 161-164, 2003

[15] Noterdaeme O and Sir Brady M, "A fast method for computing and correcting intensity inhomogeneities in MRI", IEEE Biomedical Imaging, pp. 1525-1528, 2008

[16] Moghaddam B, "Principle manifolds and probabilistic subspaces for visual recognition", IEEE Trans. PAMI 24(6), pp. 780-788, 2002

[17] Fisher, RA "The use of multiple measurements in taxonomic problems", Annals of Eugenics, pp. 179-188, 1936

[18] Fukunaga K, "Pattern recognition", Academic Press, Boston, pp. 155-157, 1990

[19] Uzunbas G, Cetin M, Unal GB, and Ercil A, "Coupled nonparametric shape prior for segmentation of multiple basal ganglia structures", 5[th] IEEE Int. Symposium on Biomedical Imaging (ISBI), pp. 217-220, May 2008

[20] Armitage P, Berry G, and Matthews JNS, "Statistical methods in medical research", Blackwell Publishing, 4[th] ed., 2002

[21] Unay D, Chen X, Ercil A, Cetin M, Jasinschi R, Buchem MA van, Ekin A, "Binary and nonbinary description of hypointensity for search and retrieval of brain MR images", SPIE Electronic Imaging, 2009

[22] Ge Y, Jensen JH, Lu H, et al, "Quantitative assessment of iron accumulation in the deep gray matter of multiple sclerosis by magnetic field correlation imaging", ANJR Am J Neuroradiol 28, pp. 1639-1644, 2007

[23] Bermel RA, Puli SR, Rudick RA, et al, "Prediction of longitudinal brain atrophy in multiple sclerosis by gray matter magnetic resonance imaging T2 hypointensity", Arch Neurol 62, pp. 1371-1376, 2005

[24] Helpern JA, Jensen J, Lee SP, et al, "Quantitative MRI assessment of Alzheimer's disease", J Mol Neurosci 24, pp. 45-48, 2004

[25] Ye FQ, Allen PS, Martin WR, "Basal ganglia iron content in Parkinson's disease measured with magnetic resonance", Movement Disorders 11, pp. 243-249, 1996

[26] Wallis LI, Paley MN, Graham JM, et al. MRI assessment of basal ganglia iron deposition in Parkinson's disease. Journal of Magnetic Resonance Imaging 28, pp. 1061-1067, 2008